# Using Wordle for Learning to Design and Compare Strategies

Chao-Lin Liu
National Chengchi University, Taiwan
chaolin@g.nccu.edu.tw

*Abstract*—Wordle has become a very popular online game since November 2021. We designed and evaluated several strategies for solving Wordle in this manuscript. Our strategies achieved impressive performances in realistic evaluations that aimed to guess all of the known answers of the current Wordle. On average, we may solve a Wordle game with about 3.67 guesses, and solve a Wordle game with six or fewer guesses higher than 98% of the time. In fact, our strategies are applicable to word guessing games that are more general than the current Wordle. More importantly, we present our work in ways that our experiences may be used as classroom examples for learning to design strategies for computer games.

## I. Introduction

The popularity of the word game **Wordle** exploded [16], and the New York Times purchased the game in 2022 [15]. Wordle is similar to *Mastermind* [11] and *Bulls and Cows* [3], but is special in that the answers are actual English words.

The main social impacts of Wordle may be in the direction of entertainment, so we found that we may utilize the popularity of this game to stimulate students' interests in designing solvers for the game from probabilistic, statistical, and information theoretical perspectives in courses like *Introduction to Artificial Intelligence* and *Computational Strategies for Games*.

The goal of playing Wordle is to find the correct answer with the least number of guesses possible. Hence, either for entertainment or for education purposes, attempts to find a theoretical solution for the minimum is not surprising, e.g., [12]. This appear to be a challenging mission mainly because Wordle has 2315 possible answers and their selection was not based on scientific reasons. The number of needed guesses will also rely on the set of correct English words that have five letters, and this set of words may not have a universal consensus on its inclusion. The distribution of the numbers of guesses needed for Wordle depends on these practical factors, so reaching a theoretical conclusion is not easy.

We may find some unofficial reports about the number of guesses that are needed for ordinary people and for programs to find the answers for Wordle. Haripriya reported the statistics that were collected on Tweeter for 241,489 games on 22 January 2022 [8]. The median and the average number of guesses were 4 and at least 4.46, respectively, where we consider the human players who failed to solve their games within six attempts would need seven guesses.

We reported a systematic method for solving the Bulls and Cows game in 2001 [2], and the method happens to be what we call "the hard mode" for Wordle today [10].[1] Playing in the hard mode sets some constraints on the guesses that a player may use, and those constraints are not necessarily easy for human to comply. Despite this obvious drawback, we re-implemented our 2001 algorithm in Python and for today's Wordle game. We randomly choose a next guess when there are multiple choices that comply the hard-mode rules. The average number of guesses used to solve the 2315 problem instances is about 4.11.

We may find that some claim to have achieved an average of 3.64 or 3.60 in personal GitHub repos or YouTube videos. The source codes were not completely open to the public for verification, so we are not sure of the reproducibility of these results.

We should like to emphasize that the main purpose of having Wordle is for entertainment. People talk about tricks and possible strategies that human beings can actually apply. Some computational methods may be more effective, but they are not proposed for human.

Building on the concept of "hard mode" which we had discussed in 2001, we invented 16 other strategies that consider probabilistic, statistical, and information-theoretical factors in the search for Wordle answers. We have found simple methods that can achieve an average of 3.85 guesses for Wordle and a relatively more computationally intensive method that can achieve an average of 3.67. We would publicize our programs for public verification along with this manuscript.

The most important contribution that we would like to make is not really about whether we offer very competitive, if not state-of-the-art, computational solutions for Wordle. Given the limited scale of Wordle from the perspective of computing powers of modern computers, one may even do exhaustive search to find a best plan for the current Wordle, e.g., [14]. That kind of success would not generalize and scale if we change the parameters for a Wordle-like game (see Section II.A for more details).

Through the discussion of our experience in designing our methods, we hope to offer some hints about the process of designing and comparing the strategies for solving computer games, and hope that the discussion can also serve as a model assignment for courses like *Introduction to Artificial Intelligence* and *Computational Strategies for Games*.

We define a class of word games that can cover the case of Wordle in Section II, where we also use a popular method as the baseline method to solve Wordle. In Sections III through V, we take a probabilistic perspective for designing strategies for Wordle, and show that a good strategy for selecting the first guesses may improve the performance of our programs. In Section VI, we adopt the ideas of learning decision trees in machine learning to design strategies, and achieved good results in the evaluation [1]. We discuss several technical issues that we experienced in this study in Section VII.

## II. Wordle and the Hard Mode

In this section, we offer a formal definition of Wordle. Our definition is more general the current Wordle games, and can

---

[1] The year is 2001, not a typo. Information about that publication is not disclosed for anonymous submission.

be used to define a family of Wordle games. We then explain how we used a "hard mode" principle to solve Wordle

*A. A Formal Defintion of the Word Game Wordle*

Let $W = (\Sigma, C, P, T, \lambda)$ denote a word game, where $\Sigma = \{s_1, s_2, \cdots, s_i, \cdots, s_n\}$, for a positive integer $n$, is a set of basic symbols. $C = \{c_1, c_2, \cdots, c_j, \cdots, c_m\}$, for a positive integer $m$, is a set of words, whereas each word $c_j \in C$ is a string of $\lambda$ symbols $c_{j1}c_{j2} \cdots c_{jk} \cdots c_{j\lambda}$ and each $c_{jk}$ is equal to a certain $s_i \in \Sigma$. A symbol may appear more than once in a word. The goal of the game is to identify the answer of the game, $T = c_a \in C$, via the shortest sequence of guesses possible. $P = \{p_1, p_2, \cdots, p_u, \cdots, p_v\}$, for a positive integer $v$, is the set of permitted words from which a player may use as a guess. To that end, a word in $P$ is string of $\lambda$ symbols in $\Sigma$, just like a word in $C$. For a reasonable game, $C$ must be a subset of $P$ or is equal to $P$.

When playing the word game, a player iteratively offers a sequence of guesses. For each guess, the player will receive a response that indicates how well the guess matches the answer. The player can choose her/his next guess according to the information that s/he infers from the previous responses in order to find $T$ with the fewest number of guesses possible.

For the Wordle game, the $\Sigma$ of Wordle is the English alphabet, all of the words in $C$ have five symbols, and $C$ is a list of 2315 different English words, i.e., $n = 26$, $m = 2315$ and $\lambda = 5$. $P$ is the set of actual English words that have exactly five letters, including some rarely used words like "CWCTH" [7], where whether a word is "actual" or not may depend on the implementation of the game providers. These settings are certainly changeable to define new games.

Let $T = t_1t_2t_3t_4t_5$ and $G = g_1g_2g_3g_4g_5$ represent the answer and a certain guess, respectively, for a Wordle game. A response $R = r_1r_2r_3r_4r_5$ to a guess consists of five squares, that can be green, yellow, or gray. A green square $r_x$ indicates that $g_x = t_x$, for $x \in \{1,2,3,4,5\}$. A yellow square $r_x$ indicates that $g_x = t_y$ for a $y \neq x$ and $x, y \in \{1,2,3,4,5\}$, on the condition that a $t_y$ can flag a $g_x$ as yellow only once. A gray square $r_x$ indicates that $g_x$ does not equal to any symbol in $T$.

*B. The Baseline Strategy: The Hard Mode*

One simple way for computers to solve Wordle is using the hard mode strategy. Assume that we have randomly chosen a first guess, $G_1$, and have received the response $R_1$. With this piece of information, we may reduce the size of $C$ with the following observation.

**Principle HM**: $c_j \in C$ cannot be the answer, if we temporarily assume $c_j$ to be the answer, use $c_j$ to compare with a guess $G_1$, and get a response that is different from $R_1$.

In the following discussion we will use 1, 2, and 0 to indicate the green, yellow, and gray square, respectively in our statements. Hence, a **perfect response** will be "11111". We will also use $C$ as $P$, although that is not necessary. We will discuss this issue in this manuscript.

The validness of the Principle HM can be explained with a simple example. If "amble" is the answer, and our guess is

---

**Strategy Hard-Mode**
Step 1. Randomly choose a $c_j$ from $C$ as the first guess $G_1$, and assume that the response is $R_1$.
Step 2. $i = 1$
Step 3. While $R_i$ is not perfect, do the following:
Step 31.  Filter and reduce $C$ with $R_i$ based on the Principle HM.
Step 32.  Randomly choose the next guess $G_{i+1}$ from the reduced $C$, and let the response be $R_{i+1}$.
Step 33   $i = i + 1$
Step 4. Record $i$. If $i > 6$, report failure.

Fig. 1. The Baseline: Strategy Hard-Mode

TABLE I. Statistics for two runs of Strategy Hard-Mode

| Strategy | min | median | mean | max |
|---|---|---|---|---|
| Hard-Mode | 1 | 4 | 4.11 | 9 |
| | excellent | failure | | |
| | 4.00% | 1.75% | | |

TABLE II. Raw records for two runs of Strategy Hard-Mode

| Number of guesses | 1 | 2 | 3 | 4 | 5 |
|---|---|---|---|---|---|
| Number of games | 3 | 182 | 1099 | 1830 | 1154 |
| Number of guesses | 6 | 7 | 8 | 9 | |
| Number of games | 281 | 60 | 20 | 1 | |

"apple", then the response is "10011". When we filter the words in $C$ with the Principle HM, we will know that "amuse" must not be the answer because, if "amuse" were the answer, we would have "10001" as the response. Hence we may exclude "amuse" from $C$ for the current game. In contrast, both "amble" and "angle" remain to be candidates for the answer.

We provide the algorithm for **Strategy Hard-Mode** that employs the Principle HM for solving Wordle in Fig. 1. The implementation is really easy, and the computation is very efficient. Tables I and II show the statistics of two runs of Strategy Hard-Mode on the 2315 Wordle answers. Since the guesses were randomly selected, we could observe different outcomes in repeated runs. Since we conduct the experiments twice, the total number of games is 4630 in Table II, and we solved 1830 games with four guesses. On average, we used 4.11 guesses to solve the games, and failed to find the answers with six or fewer attempts in (60+20+1)=81 games, which is equivalent to 1.75% "failure" rate. We considered games in which we found the answers with one or two guesses as "excellent". The baseline methods performed excellently in 4.00% of the games.

### III. COLLOCATION-BASED HEURISTIC

After using the Principal HM to filter $C$, we have used up the information that the responses to the previous guesses have offered. All of the words in the reduced $C$ are reasonable candidates for the answer. In the Strategy Hard-Mode, a word in $C$ was chosen as the next guess randomly. We need to invent a heuristic to choose the next guess from the reduced $C$.

*A. Motivation*

Recall the game of aiming to guess a number between 1 and 10.[2] The optimal strategy is using the current guess to split the remaining candidates of the answer into two subgroups of almost equal sizes each time. By doing so, we minimize the depth of the search tree, and minimize the expected number of steps needed to find the answer.

---

[2] http://www.learningaboutelectronics.com/Articles/Number-guessing-game-with-PHP.php

This observation also provides a motivation for understanding the design of binary search tree [5]. We may inherit the ideas of binary search trees, and estimate the quality of the groupings of the remaining answers in $C$ of the Wordle based on the unconditional and conditional distributions of the symbols.

*B. Unconditional and Conditional Probability of Symbols*

For any given $C$ and $P$ in a game, it is easy for us to compute the unconditional and conditional probabilities of inclusion of the symbols in words.

We define the unconditional probability of a symbol $s_i$ in $\Sigma$ for the $C$ as the probability of the inclusion of $s_i$ in the words in $C$. The unconditional probability of a symbol $s_i$ in $\Sigma$ for the $P$ is defined analogously. Identity (1) provides an operational definition for $Pr_c(s_i)$.

$$Pr_c(s_i) = \frac{number\ of\ words\ that\ include\ s_i\ in\ C}{number\ of\ words\ in\ C} \quad (1)$$

We define the conditional probability $Pr_c(x|s_i)$ of seeing a symbol $x$ given that the symbol $s_i$ is present in a word in $C$. The conditional probability for $P$ is defined analogously. Identity (2) provides an operational definition for $Pr_c(x|s_i)$ for all symbols $x$ in $\Sigma$, where $Pr_c(x, s_i)$ is the probability that $x$ and $s_i$ appear in the same word in $C$. $Pr_c(s_i|s_i)$ may not be zero if there are words in $C$ that include more than one $s_i$.

$$Pr_c(x|s_i) = \frac{Pr_c(x, s_i)}{Pr_c(s_i)} \quad (2)$$

Given the conditional probability values for all symbols in $\Sigma$, we may compute the entropy for each of these conditional distributions $H_c(s_i)$, using the identity shown in (3).

$$H_c(s_i) = \sum_{k=1}^{k=n} Pr_c(s_k|s_i) \log \frac{1}{Pr_c(s_k|s_i)} \quad (3)$$

*C. Ranking the Candidates Words*

The task of selecting the next word as our guess requires us to compute a score for a candidate word in $C$. Recall that, in the process of playing Wordle, the size of $C$ decreases in each iteration in the Strategy Hard-Mode, so the computation of the unconditional probability, condition probability, and the entropy is a dynamic task.

Each word in a word game $G$ has $\lambda$ symbols. If we naively assume that the contributions of each of these $\lambda$ symbols to the score of a candidate word are independent, we have a simple way to estimate the score of the candidate words in $G$. This step is expressed in identity (4).

$$\text{score}(c_j) = \sum_{k=1}^{k=\lambda} score(c_{jk}) \quad (4)$$

*D. Maximizing the Entropy when Ranking the Candidates*

From here, we have multiple ways to define $score(c_{jk})$. Some of which are intuitively favorable, and others appear to be less attractive. In a university course, students may be encouraged to try and compare their actual effects.

Based on the nature of the entropy, if we prefer the $s_i \in \Sigma$ that has a larger $H_c(s_i)$, we are favoring the $s_i$ that collocates more diversely with the symbols in $\Sigma$. Getting information about such an $s_i$ allows us to collect more information about more symbols in $\Sigma$, therefore increasing the possibility of leading to more shallow search tree. The score for a candidate word can be as simple as identity (5) shows, if we continue to choose the next guess from the current reduced as we explained in Section II.B.

$$\text{score}(c_j) = \sum_{k=1}^{k=\lambda} score(c_{jk}) = \sum_{k=1}^{k=\lambda} H_c(c_{jk}) \quad (5)$$

In (5), the contribution of a symbol $c_{jk}$ in $c_j$ is the entropy of its collocational probability, by setting $s_i = c_{jk}$ in (2) and (3).

TABLE III. Statistics for Strategy Hard-Mode-Collocation

| Strategy | un-max | un-min | wht-max | wht-min |
|---|---|---|---|---|
| min | 1 | 1 | 1 | 1 |
| median | 4 | 5 | 5 | 5 |
| mean | 4.326 | 5.044 | 4.62 | 4.525 |
| max | 11 | 10 | 10 | 9 |
| excellent | 2.59% | 1.47% | 2.42% | 2.38% |
| failure | 4.71% | 10.58% | 7.65% | 3.24% |

It is intriguing to weigh $H_c(c_{jk})$ by the unconditional probability of $Pr_c(c_{jk})$ when calculating $score(c_{jk})$. Identity (6) shows the operation for this intuitive exploration.

$$\text{score}(c_j) = \sum_{k=1}^{k=\lambda} score(c_{jk}) = \sum_{k=1}^{k=\lambda} \frac{Pr_c(c_{jk}) H_c(c_{jk})}{normalizer}, \quad (6)$$

where $normalizer = \sum_{k=1}^{k=\lambda} Pr_c(c_{jk})$

Putting the above reasoning together we would choose the $c_j$ that has the largest $score(c_j)$ for all current candidate words. This step is expressed in the following identity

$$\text{nextGuess}(C) = c_j^* = argmax_{c_j \in C} score(c_j) \quad (7)$$

Recall that our using *argmax* in (7) is based on intuitive arguments. It is thus educational to switch to using *argmin* in part of our evaluation process. We may examine whether or not results of realistic experiments support our intuition. For this purpose, we used the identity in (8).

$$\text{nextGuess}(C) = c_j^* = argmin_{c_j \in C} score(c_j) \quad (8)$$

*E. Algorithm and its Evaluation*

We replace the steps of randomly selecting the next guess in Strategy Hard-Mode in Fig. 1 with the steps that aim to optimize either (7) or (8), depending on the goals of individual experiments. Fig. 2 shows the algorithm for **Strategy Hard-Mode-Collocation**.

Table III shows the summary for the experiments in which we may use four possible different ways to choose the next guesses. The label un-max (for *unweighted-argmax*), indicate that identities (5) and (7) were used in the experiment, un-min (for *unweighted-argmin*) indicates that (5) and (8) were used, wht-max (for *weighted-argmax*) indicates that (6) and (7) were used, and wht-min (for *weighted-argmin*) indicates that (6) and (8) were used.

```
Strategy Hard-Mode-Collocation
Step 1. Choose the $c_j$ from $C$ that optimize score$(c_j)$, $c_j \in C$,
        based on the identities (7) or (8), as the first guess $G_1$,
        and assume that the response is $R_1$.
Step 2. $i = 1$
Step 3. While $R_i$ is not perfect, do the following:
Step 31.   Filter and reduce $C$ with $R_i$ based on the Principle
           HM.
Step 32.   Choose the next guess $G_{i+1} = c_j^*$ whose score is
           maximum among the candidates in the reduced $C$,
           and let the response be $R_{i+1}$. Again, we may use
           identities (7) or (8) at this step.
Step 33    $i = i + 1$
Step 4. Record $i$. If $i > 6$, report failure.
```

Fig. 2. The Baseline: Strategy Hard-Mode

It was quite disappointing that none of these strategies outperformed the baseline strategy, at initially. We found that Strategy Hard-Mode-Collocation tended to choose words with repeated characters for the first and may be for the second guesses. Words like "fuzzy", "vivid", and "knock" were common.

IV. THE POLICTY ON SELECTING THE FIRST GUESSES

Gradually, we consider more heuristics to improve our algorithms. When selecting the next guesses with (7) or (8),

TABLE IV. Statistics for Strategy Hard-Mode-Collocation with constraints on selecting first guesses

| Strategy | un-max | un-min | wht-max | wht-min |
|---|---|---|---|---|
| min | 1 | 1 | 1 | 1 |
| median | 4 | 5 | 4 | 4 |
| mean | 3.906 | 4.674 | 4.551 | 4.245 |
| max | 9 | 9 | 11 | 9 |
| excellent | 5.36% | 2.29% | 3.54% | 3.63% |
| failure | 2.07% | 5.57% | 8.16% | 1.68% |

we do not consider the distributions of the symbols that form the words. Hence, it is possible that a symbol might appear more than once in competitive candidate words. Having repeated symbols in a guess is particularly unattractive, at least intuitively, for the very first guess in Wordle. One possible and common policy is to select words that do not have repeated symbols at least for the first guess. Among the 2315 possible answers for Wordle, 1655 words do not have repeated symbols.

We added this constraint to the Strategy Hard-Mode-Collocation, and re-ran our experiments. Table IV shows the results. The performances improved across the board when we compare the corresponding items in Tables III and IV, except that the results of using *weighted-argmin* improved only partially.

It is worthwhile mentioning that the results shown in the un-max column in Table IV are better than their corresponding items listed in Table I. The average number of guesses needed to find the answers was reduced, the excellent rate was increased, and the failure rate was reduced.

The differences in the performance metrics between *unweighted-argmax* and *unweighted-argmin* supported our reasoning for using identities (5) and (7). The negative impacts of replacing (7) with (8) were salient. We tried

TABLE V. Statistics for the Strategy Hard-Mode-Collocation-KLD

| Strategy | min | median | mean | max |
|---|---|---|---|---|
| Hard-Mode-Collocation-KLD | 1 | 4 | 3.851 | 10 |
|  | excellent | failure | | |
|  | 5.75% | 1.73% | | |

TABLE VI. Raw records for the Strategy Hard-Mode-Collocation-KLD

| Number of guesses | 1 | 2 | 3 | 4 | 5 |
|---|---|---|---|---|---|
| Number of games | 1 | 132 | 1099 | 910 | 355 |
| Number of guesses | 6 | 7 | 8 | 9 | 10 |
| Number of games | 103 | 29 | 7 | 3 | 1 |

*weighted-argmax* and *weighted-argmin* just because of curiosity, and their performances were poorer than those of the baseline method.

V. INFORMATION-THEORETIC APPROACHS

We also tried to apply the concept of the Kullback-Leibler divergence between the conditional probability distribution $Pr_c(s_k|s_i)$ and the discrete uniform distribution that assumes that all $s_k$ are equally likely [9]. Therefore, we can carry out a simple derivation that is provided in the Appendix. Preferring an $s_i$ that has larger $score(s_i)$ in identity (9) is tentative to favoring a conditional probability that is more different from a uniform distribution. This might sound like a reasonable factor for a good guess, but the rationality is not very strong. Despite this vagueness, we replaced identity (3) in Section III.B with (9), and named the new strategy Hard-Mode-Collocation-KLD. The experimental results, listed in Table V, are close to and better than those listed in the un-max column in Table IV. Table VI lists the actual distribution of the numbers of guesses that we used to solve the 2315 problems.

$$score(s_i) = \sum_{k=1}^{k=n} Pr_c(s_k|s_i) \log(n Pr_c(s_k|s_i)) \quad (9)$$

VI. HIGHER-LEVEL SEARCH CONSIDERATIONS

Assume that we are working on Wordle games whose answers are 5-letter words, and that we have chosen a word $c_j$ in $C$ as a guess. The response must be one of the patterns listed in Table VII. Therefore, we many consider that a guess would lead us to cluster the words into 20 groups, and members of each of these groups would give our guess the same response that is specific for that group. It should be easy to understand that if the answers for Wordle have more number of letters, it would be time consuming to make a table like TABLE VII manually, but that is doable computationally. From this perspective, we may say that a guess will divide the current reduced $C$ into sub-sets.

Due to this observation, we can calculate the percentages of the words in the sub-sets, and use the percentages as a probability distribution to calculate the resulting entropy when we use a guess to divide current $C$. Analogous to our trying to maximizing the Information Gain when we build decision trees in machine learning, we would prefer to minimize the resulting entropy when we use a guess to divide the current $C$. Moreover, we may employ the concept of the Kullback-Leibler divergence to compute the scores of choosing a certain candidate word for Wordle. The process is similar to the development that we discussed in details in Sections III, IV, and V. Although the process is similar, the computation procedures are much more time consuming than using the collocation-based information.

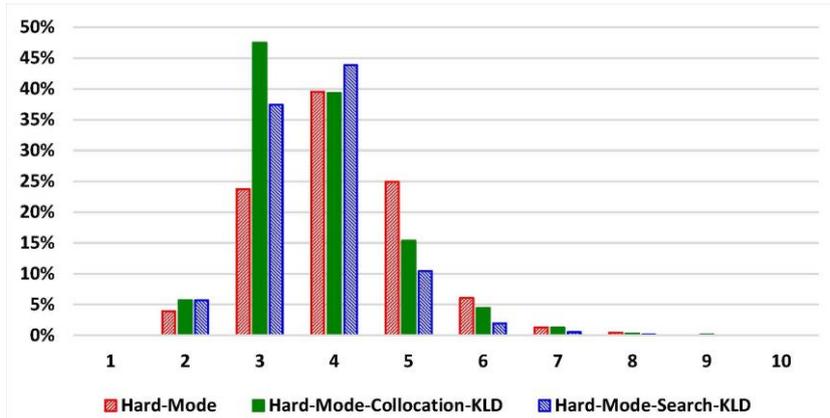

Fig. 3. Distributions of the percentages of numbers of guesses used to solve the 2315 Wordle games

TABLE VII. Possible Responses of Wordle (5-letter words)

| ID | number of green squares | number of yellow squares | number of gray squares |
|---|---|---|---|
| 1 | 5 | 0 | 0 |
| 2 | 4 | 0 | 1 |
| 3 | 3 | 2 | 0 |
| 4 | 3 | 1 | 1 |
| 5 | 3 | 0 | 2 |
| 6 | 2 | 3 | 0 |
| 7 | 2 | 2 | 1 |
| 8 | 2 | 1 | 2 |
| 9 | 2 | 0 | 3 |
| 10 | 1 | 4 | 0 |
| 11 | 1 | 3 | 1 |
| 12 | 1 | 2 | 2 |
| 13 | 1 | 1 | 3 |
| 14 | 1 | 0 | 4 |
| 15 | 0 | 5 | 0 |
| 16 | 0 | 4 | 1 |
| 17 | 0 | 3 | 2 |
| 18 | 0 | 2 | 3 |
| 19 | 0 | 1 | 4 |
| 20 | 0 | 0 | 5 |

More specifically, let $\Gamma = \{\gamma_1, \gamma_2, \cdots, \gamma_a, \cdots, \gamma_b\}$ denote the set of all possible responses for a word game $W$. Table VII shows the $\Gamma$ for a Wordle game whose answers are words of five symbols. We may conceptually divide the current $C$ of $W$ into $b$ groups, with a guess $G$ as following: If $c_j \in C$ is the answer of $W$ and if its response to $G$ is $\gamma_a$, then we put $c_j$ into the group $g(\gamma_a)$. Therefore, by construction, each word in must belong to a certain group in $\Gamma$.

TABLE VIII. Statistics for the Strategy Hard-Mode-Search-KLD

| Strategy | min | median | mean | max |
|---|---|---|---|---|
| Hard-Mode-Search-KLD | 1 | 4 | 3.674 | 8 |
| | excellent | failure | | |
| | 5.75% | 0.65% | | |

TABLE IX. Raw records for the Strategy Hard-Mode-Search-KLD

| Number of guesses | 1 | 2 | 3 | 4 | 5 |
|---|---|---|---|---|---|
| Number of games | 1 | 132 | 866 | 1015 | 241 |
| Number of guesses | 6 | 7 | 8 | | |
| Number of games | 45 | 12 | 3 | | |

We can define a probability distribution based on the memberships of these groups. Let $s(\gamma_a)$ be the number of words in $g(\gamma_a)$. Hence, if there are $x$ words in the current $C$, the following identity must hold.

$$\sum_{a=1}^{b} s(\gamma_a) = x \qquad (10)$$

Therefore, let $p(\gamma_a) = \frac{s(\gamma_a)}{x}$, and we have the following.

$$\sum_{a=1}^{b} p(\gamma_a) = 1 \qquad (11)$$

With these basic setups, we can define the resulting entropy and Kullback-Leibler divergence in ways that are very similar to what we reported in Section III, IV, and V, when we choose a guess, $G$, to divide the current $C$. We can then use the entropy and the divergence to compare candidate guesses and choose our next guess, to enhance the baseline Strategy Hard-Mode and establish the Hard-Mode-Search-KLD strategy.

Tables VIII and IX lists the best results that this relatively more complex procedure could achieve. This Hard-Mode-Search-KLD strategy led to slightly better performance, i.e., the average and maximal numbers of guesses to solve the game and the failure rates were improved. The distributions recorded in Tables IX and VI are quite different.

Fig. 3 depict the distributions in percentages for the data in Tables II, VI, and IX. Our introducing different methods to choose the first guess and the next guesses for a Wordle game paid off. Using the Hard-Mode-Collocation-KLD and the Hard-Mode-Search-KLD strategies, we were more likely to find the answers with three or fewer guesses, while reducing the possibility of needing five or more guesses to solve the games. The proportion of excellent games increased, and the proportion of failed games decreased.

## VII. DISCUSSION

We have used the Hard-Mode Strategy as the baseline. The strategy performs pretty well in practice, c.f. Tables I and II. We have found and Peattle also discussed that this strategy may not work well for some special cases [12].

Assume that the answer is "freed", that we have guessed "creed", and that we got the response of [gray, green, green, green, green]. In this case, if playing in the Hard Mode, we may have to try "greed" and "breed" before we find the correct answer. An even more challenging group of words include "goner", "cover", "wooer", "homer", "poker", and "foyer". Allowing not to abide by the hard-mode rules sometimes will help. It may not be easy to find an answer in the group "wight", "fight", "sight", "tight", "right", "night", "light", and "eight" with no more than six attempts under the hard-mode rules.

For simplifying our discussion, we have used $C$ as $P$. In practice, there are a lot more words in $P$ than in $C$. It is easy to find good resources about English words that have five letters online, e.g., [4]. Using $C$ as $P$ is not a required trick for our programs. On one hand, using words in $C$ as our guesses gave us some chances to directly find the answers luckily. On the other hand, we also wonder whether using a word in $P$ will provide more information than using any other words in $C$.

We have mentioned that we consider that a main contribution of this manuscript is to provide the experience in developing strategies for solving a class of word games. The word game $W$ as we defined in Section II.A is flexible, and one may change the parameters as long as one wish. For instance, the words for answers may not have to be English words, and it is possible for one to define games that include more symbols than the English alphabet in $\Sigma$.

We hope that the examples of our designing and choosing the heuristics to guide the selection of next guesses may be used as classroom examples of designing and comparing strategies for computer games.

We evaluated our methods with a single Wordle so far. One may apply our methods to solve Dordle [6] and Quordle [13] in which a plyer needs to solve more than one Wordle at a time. If each of these Wordle games are independent, then our methods should be directly applicable. If individual Wordle games are dependent, it should be possible to enhance our current design to handle the extra constraints.

## VIII. CONCLUDING REMARKS

We have proposed a few strategies for solving a special class of word games, and used typical Wordle games as the example problems. Results of realistic evaluation indicate that we have achieved competitive performances for the current Wordle. In addition to providing clues for solving Wordle, we hope that the process of inventing and evaluating candidate strategies could serve as classroom examples for courses on learning to design strategies for computer games.

## ACKNOWLEDGMENT

This work was funded in part by the MOST-110-2221-E-004-008-MY3 project of the Ministry of Science and Technology of Taiwan and in part by the 111H124D-13 project of the National Chengchi University.

## REFERENCES

[1] E. Alpaydin, Introduction to Machine Learning, fourth edition, MIT Press, 2020.

[2] C.-L. Liu. Mathematics, computer science, and number games (數學、資訊科學與數字遊戲), *Science Monthly* (科學月刊) 32(3), 250–255, 2001. (in Chinese)

[3] Bulls and Cows: https://en.wikipedia.org/wiki/Bulls_and_Cows

[4] CMUDict: https://pypi.org/project/cmudict/

[5] T. H. Cormen, C. E. Leiserson, R. L. Rivest, and C. Stein, Introduction to Algorithms, chapter 12, the third edition, MIT Press, 2009.

[6] Dordle: https://zaratustra.itch.io/dordle

[7] GAMERANT, https://gamerant.com/wordle-words-with-no-vowels/

[8] Haripriya, "What are the average number of guesses in Wordle?", https://nerdschalk.com/what-are-the-average-number-of-guesses-in-wordle/

[9] S. Kullback and R.A. Leibler, "On information and sufficiency," Annals of Mathematical Statistics, 22 (1): 79–86, 1951.

[10] L. Loofbourow and M. Martinelli, "Should you be playing Wordle on "Hard Mode"?", SLATE, https://slate.com/culture/2022/02/wordle-game-nyt-original-vs-hard-mode.html

[11] Mastermind: https://en.wikipedia.org/wiki/Mastermind_(board_game)

[12] A. Peattle, "Establishing the minimum number of guesses needed to (always) win Wordle," personal blog, https://alexpeattie.com/blog/establishing-minimum-guesses-wordle/

[13] Quordle: https://www.quordle.com/#/

[14] A. Selby, "The best strategies for Wordle," http://sonorouschocolate.com/notes/index.php?title=The_best_strategies_for_Wordle

[15] M. Tracy, "The New York Times buys Wordle," New York Times, 31 Jan 2022, https://www.nytimes.com/2022/01/31/business/media/new-york-times-wordle.html

[16] Wordle: https://en.wikipedia.org/wiki/Wordle

## APPENDIX

In the following derivation, $U$ denote a uniform distribution that we want to compare with the conditional probability distribution $Pr_c(s_k|s_i)$, for a specific $s_i$. Since $s_k$ can be any symbol in $\Sigma$, we need a uniform random variable that could take the value of any state among $|\Sigma|$ states. Since $\Sigma = \{s_1, s_2, \cdots, s_i, \cdots, s_n\}$, we have $|\Sigma| = n$.

$$\begin{aligned}
score(s_i) &= KLD_c(Pr_c(s_k|s_i) \| U) \\
&= \sum_{k=1}^{k=n} Pr_c(s_k|s_i) \log \frac{Pr_c(s_k|s_i)}{\left(\frac{1}{|\Sigma|}\right)} \\
&= \sum_{k=1}^{k=n} Pr_c(s_k|s_i) \log(n \, Pr_c(s_k|s_i))
\end{aligned}$$

Table A1 lists the statistics of the outcomes, including the minimum, median, average, and the maximum of the numbers of guesses that were used by different strategies to solve the 2315 problems. The excellency column shows the percentages of a strategy using only one or two guesses to solve the 2315 problems. The failure column shows the percentages of a strategy using seven or more guesses to solve the 2315 problems.

Figure A1 depicts the distributions of the numbers of guesses that were used by different strategies. We show the strategies at the bottom, where the "hard-mode" is the baseline, the "i" and "p" families of strategies were denoted by "i" and "p" that were followed by a digit, respectively. We show the number of needed guesses on the horizontal axis, and the frequencies of the number of needed guesses on the vertical axis.

**Table A1. Basic statistics**

| strategy | min | median | mean | maximum | excellency | failure |
|---|---|---|---|---|---|---|
| hard-mode | 1 | 4 | 4.078 | 10 | 4.67% | 1.77% |
| i1 | 1 | 6 | 5.651 | 11 | 1.47% | 28.51% |
| i2 | 1 | 4 | 4.117 | 9 | 2.59% | 1.34% |
| i3 | 1 | 4 | 4.475 | 10 | 2.59% | 5.49% |
| i4 | 1 | 4 | 3.674 | 8 | 5.75% | 0.65% |
| i5 | 1 | 5 | 4.926 | 10 | 2.29% | 11.27% |
| i6 | 1 | 4 | 3.750 | 9 | 5.75% | 0.52% |
| i7 | 1 | 4 | 4.205 | 9 | 3.20% | 2.98% |
| i8 | 1 | 4 | 3.674 | 8 | 5.75% | 0.65% |
| p1 | 1 | 4 | 4.263 | 10 | 2.72% | 2.76% |
| p2 | 1 | 4 | 4.301 | 10 | 2.72% | 3.11% |
| p3 | 1 | 5 | 4.525 | 9 | 2.38% | 3.24% |
| p4 | 1 | 5 | 4.583 | 9 | 2.38% | 3.41% |
| p5 | 1 | 4 | 3.851 | 10 | 5.75% | 1.73% |
| p6 | 1 | 4 | 3.848 | 10 | 5.75% | 1.56% |
| p7 | 1 | 4 | 4.245 | 9 | 3.63% | 1.68% |
| p8 | 1 | 4 | 4.236 | 9 | 3.63% | 1.47% |

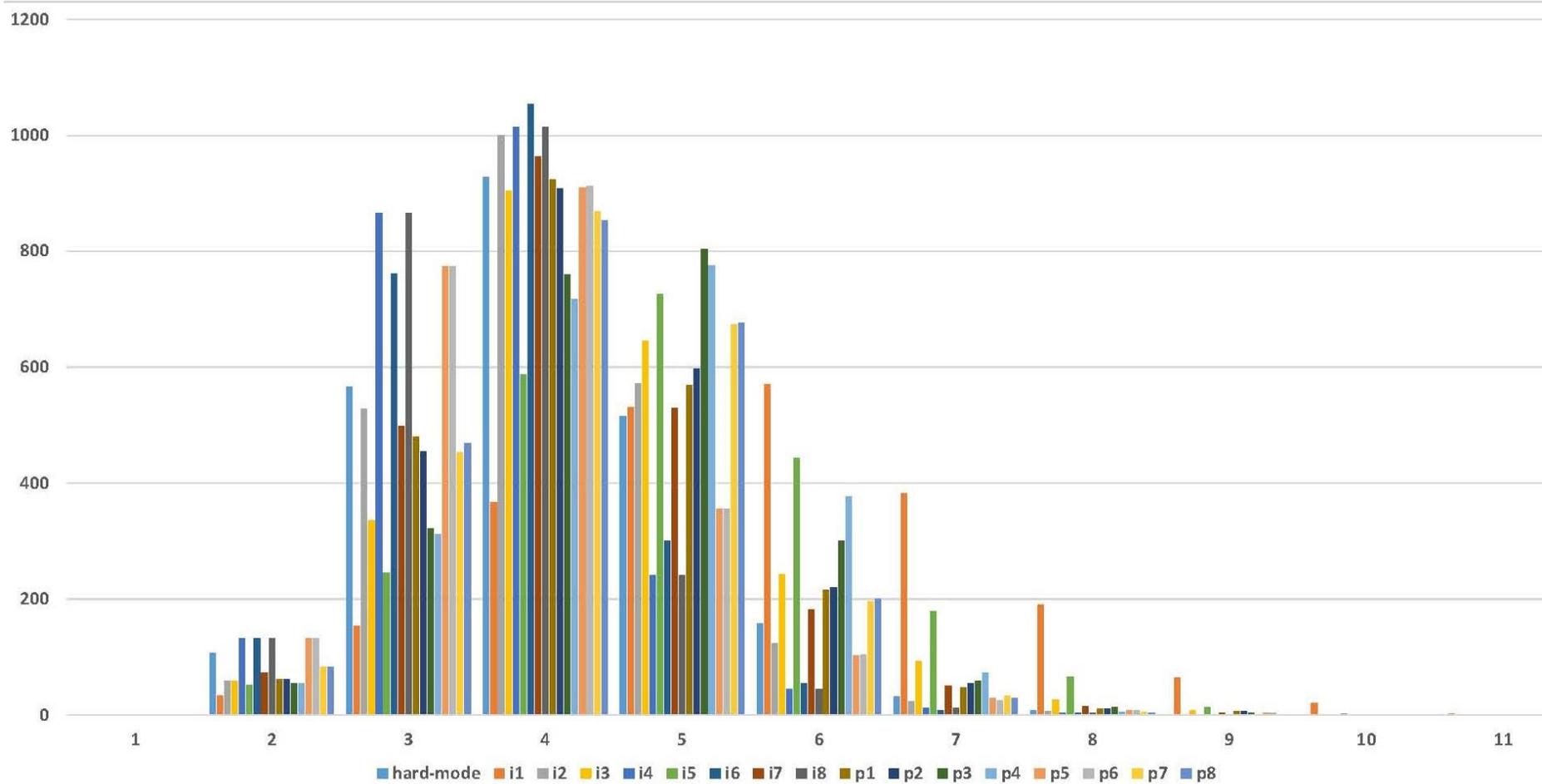

Figure A1. Distributions of the numbers of guesses that were used by different strategies to solve the 2315 problems